# Natural Language Processing for EHR-Based Computational Phenotyping

Zexian Zeng, Yu Deng, Xiaoyu Li, Tristan Naumann, Yuan Luo

**Abstract**— This article reviews recent advances in applying natural language processing (NLP) to Electronic Health Records (EHRs) for computational phenotyping. NLP-based computational phenotyping has numerous applications including diagnosis categorization, novel phenotype discovery, clinical trial screening, pharmacogenomics, drug-drug interaction (DDI) and adverse drug event (ADE) detection, as well as genome-wide and phenome-wide association studies. Significant progress has been made in algorithm development and resource construction for computational phenotyping. Among the surveyed methods, well-designed keyword search and rule-based systems often achieve good performance. However, the construction of keyword and rule lists requires significant manual effort, which is difficult to scale. Supervised machine learning models have been favored because they are capable of acquiring both classification patterns and structures from data. Recently, deep learning and unsupervised learning have received growing attention, with the former favored for its performance and the latter for its ability to find novel phenotypes. Integrating heterogeneous data sources have become increasingly important and have shown promise in improving model performance. Often better performance is achieved by combining multiple modalities of information. Despite these many advances, challenges and opportunities remain for NLP-based computational phenotyping, including better model interpretability and generalizability, and proper characterization of feature relations in clinical narratives.

——————————   ◆   ——————————

## 1   INTRODUCTION

A phenotype is an expression of the characteristics that result from genotype variations and an organism's interactions with its environment. A phenotype may consist of physical appearances (e.g., height, weight, BMI), biochemical processes, or behaviors [1]. In the medical domain, phenotypes are often summarized by experts on the basis of clinical observations. Nationwide adoption of Electronic Health Records (EHRs) has given rise to a large amount of digital health data, which can be used for secondary analysis [2]. Typical EHRs include structured data such as diagnosis codes, vitals and physiologic measurements, as well as unstructured clinical narratives such as progress notes and discharge summaries. Computational phenotyping aims to automatically mine or predict clinically significant, or scientifically meaningful, phenotypes from structured EHR data, unstructured clinical narratives, or their combination.

As summarized in a 2013 review by Shivade et al. [3], early computational phenotyping studies were often formulated as supervised learning problems wherein a predefined phenotype is provided, and the task is to construct a patient cohort matching the definition's criteria. Many of these studies relied heavily on structured and coded patient data; for example, using encodings such as International Classification of Disease, 9th Revision (ICD-9) [4], its successor the 10th Revision (ICD-10) [5], Systematized Nomenclature of Medicine-Clinical Terms (SNOMED CT) [6], RxNorm [7], and Logical Observation Identifiers Names and Codes (LOINC) [8]. On the other hand, the use of natural language processing (NLP) for EHR-based computational phenotyping has been limited to term and keyword extraction [3].

Structured data typically capture patients' demographic information, lab values, medications, diagnoses, and encounters [9]. Although readily available and easily accessible, studies have concluded that structured data alone are not sufficient to accurately infer phenotypes [10, 11]. For example, ICD-9 codes are mainly recorded for administrative purposes and are influenced by billing requirements and avoidance of liability [12, 13]. Consequently, these codes do not always accurately reflect a patient's underlying physiology. Furthermore, not all patient information is well documented in structured data, such as clinicians' observations and insights [14]. As a result, using structured data alone for phenotype identification often results in low performance [11]. The limitations associated with structured data for computational phenotyping have encouraged the use of clinical narratives, which typically include clinicians' notes, observations, referring letters, specialists' reports, discharge summaries, and a record of communications between doctors and patients [15]. Unstructured clinical narratives may summarize patients' medical history, diagnoses, medications, immunizations, allergies, radiology images, and laboratory test results, in the forms of progress notes, discharge notes etc. [16].

Structured and unstructured EHR data are often stored in vendor applications or at a healthcare enterprise data warehouse. Typical

——————————————

- *Zexian Zeng is with the Department of Preventive Medicine, Northwestern University Feinberg School of Medicine, Chicago, IL 60611. E-mail: zexian.zeng@northwestern.edu.*
- *Yu Deng is with the Department of Preventive Medicine, Northwestern University Feinberg School of Medicine, Chicago, IL 60611. E-mail: yu.deng@northwestern.edu.*
- *Xiaoyu Li is with the Department of Social and Behavioral Sciences, Harvard T.H. Chan School of Public Health, Boston, MA 02115. E-mail: xil288@mail.harvard.edu.*
- *Tristan Naumann is with Computer Science and Artificial Intelligence Lab, Massachusetts Institue of Technology, Cambridge, MA 02139. E-mail: tjn@mit.edu.*
- *Yuan Luo is with the Department of Preventive Medicine, Northwestern University Feinberg School of Medicine, Chicago, IL 60611. E-mail: yuan.luo@northwestern.edu.*

EHR data are usually managed by a local institution's technicians and are accessible to trained personnel or researchers. Institutional Review Boards at local institutions typically grant access to certain patient cohorts and certain parts of EHRs. Database queries can then be written and executed to retrieve desired structured and unstructured EHR data. In addition to hospital-collected data stored in EHRs, research data are increasingly available, including public databases such as PubMed [17], Textpresso [18], Human Protein Interaction Database (HPID) [19], and MeInfoText [20]. With growing amount of available data, efficient identification of relevant documents is essential to the research community. Information retrieval systems have been developed to identify text corresponding to certain topics or areas from EHR data across multiple fields. CoPub Mapper [21] ranks co-occurrence associations between genes and biological terms from PubMed. iHOP [22] links interacting proteins to their corresponding databases and uses co-occurrence information to build a graphical interaction network. We refer the reader to the following reviews for more details: [23] is a survey for biomedical text mining in cancer research, [24] is a survey for biomedical text mining, and [25] is a survey for web mining.

While the prevalence of EHR data presents an opportunity for improved computational phenotyping, extracting information from clinical narratives for accurate phenotyping requires both semantic and syntactic structures in the narrative to be captured [26]. Scaling such tasks to large cohort studies is laborious, time-consuming, and typically requires extensive data collection and annotation.

Recently, NLP methods for EHR-based computational phenotyping have seen extensive development, extending beyond basic term and keyword extraction. One focus of recent studies is formulating computational phenotyping as an unsupervised learning problem to automatically discover unknown phenotypes. The construction of richer features such as relations between medical concepts enables greater expressive power when encoding patient status, compared to terms and keywords. More advanced machine learning methods, such as deep learning, have also been increasingly adopted to learn the underlying patient representation.

This article reviews the literature on NLP methods for EHR-based computational phenotyping, emphasizing recent developments. We first describe several applications of computational phenotyping. We then summarize the state-of-the-art NLP methods for computational phenotyping and compare their advantages and disadvantages. We also describe the combinations of data modalities, feature learning, and relation extraction that have been used to aid computational phenotyping. Finally, we discuss challenges and opportunities to NLP methods for computational phenotyping and highlight a few promising future directions.

# 2 APPLICATIONS OF EHR-BASED COMPUTATIONAL PHENOTYPING

Computational phenotyping has facilitated biomedical and clinical research across many applications, including patient diagnosis categorization, novel phenotype discovery, clinical trial screening, pharmacogenomics, drug-drug interaction (DDI) and adverse drug event (ADE) detection, and downstream genomics studies.

## 2.1 Diagnosis Categorization

One of the most important applications of computational phenotyping is diagnosis categorization, which enables the automated and efficient identification of patient cohorts for secondary analysis [15, 27-31]. A wide range of diseases has been investigated in the past, including suspected tuberculosis (TB) [32, 33], colorectal cancer [34], rheumatoid arthritis [35], diabetes [36], heart failure [37, 38], neuropsychiatric disorders [39], etc. These applications have extended from disease identification to disease subtyping such as lung cancer stage evaluation [40], or subsequent event detection such as breast cancer recurrence detection [41] and cancer metastases detection [42].

## 2.2 Novel Phenotype Discovery

Computational phenotyping has been applied to discover novel phenotypes and sub-phenotypes. Traditionally, a clinical phenotype is classified into a particular category if it meets a set of criteria developed by domain experts [43]. Instead, semi-supervised or unsupervised methods can detect traits based on intrinsic data patterns with moderate or minimal expert guidance, which may promote the discovery of novel phenotypes or sub-phenotypes. For example, in a study by Marlin et al. [44], a diagonal covariance Gaussian mixture model was applied on physiological time series data for patient clustering. They discovered distinct, recognizable physiological patterns and they concluded that interpretations of these patterns could offer prognostic significance. Doshi-Velez et al. [45] applied hierarchical clustering to define subgroups with distinct courses among autism spectrum disorders. They applied ICD-9 codes to construct time series features. In the study, they identified four subgroups among 4934 patients; one subgroup was characterized by seizures; one subgroup was characterized by multisystem disorders including gastrointestinal disorders, auditory disorders, and infections; one subgroup was characterized by psychiatric disorders; one subgroup could not be further resolved. In a study by Ho et al. [46], they applied tensor factorization [47, 48] on medication orders to generate phenotypes without supervision. In a case study searching for 50 phenotypes in heart failure, they achieved better performance than principal component analysis (PCA) with respect to area under curve (AUC) score and model stability. Further interpretations of these novel phenotypes have potential to offer us useful clinical information. Shah et al. [49] clustered patients with preserved ejection fraction into three novel subgroups, which offers meaningful insight into clinical characteristics, cardiac structures, and outcomes.

## 2.3 Clinical Trial Screening

Leveraging EHR data can benefit clinical trial recruitment [50]. In recent years, echoing the rising availability of EHR data and the increased volume of clinical trial recruitments, computational phenotyping for clinical trial screening has become an active area. Multiple systems have been designed for this purpose [51-54]. Electronic screening can improve efficiency in clinical trial recruitment, and automated querying over trials can support clinical knowledge curation [55]. A typical computational phenotyping system for clinical trial eligibility identifies patients whose profiles—extracted from structured data and narratives—matched the trial criteria in order to reduce the pool of candidates for further staff screening.

## 2.4 Pharmacogenomics

Pharmacogenomics aims to investigate the interaction between genes, gene products, and therapeutic substances. Much of this knowledge exists in scientific literature and curated databases. Computational phenotyping applications have been developed to mine pharmacogenomics knowledge [56-59]. These phenotyping tools automatically scan, retrieve, and summarize the literature for meaningful phenotypes. Recent studies have adopted semantic and syntactic analyses as well as statistical machine learning tools to mine targeted pharmacogenomics relations from scientific literature and clinical records [58].

## 2.5 DDIs and ADEs

Drug-drug interactions (DDIs) happen when one drug affects the activity of another drug that has been simultaneously administered. Adverse drug events (ADEs) refer to unexpected injuries caused by administering medication. Detecting DDIs and ADEs can guide the process of drug development and drug administration. The impact of these negative outcomes has triggered huge efforts from industry and the scientific community to develop models exploring the relationships between drugs and biochemical pathways in order to enable the discovery of DDIs [60, 61] and ADEs [26, 62-64].

## 2.6 GWAS and PheWAS

Cohorts obtained by computational phenotyping have benefited downstream genomic studies [65], using techniques such as Genome-wide association studies (GWAS) and phenome-wide association studies (PheWAS). In GWAS, researchers link genomic information from DNA biorepositories to EHR data to detect associations between phenotypes and genes. In such studies, case-control cohorts can be generated without labor intensive annotation, which is especially important for rare variant studies where a large number of patients need to be screened. Much research [66-69] has explored EHR phenotyping algorithms to facilitate GWAS. We refer the reader to reviews by Bush et al. [70] and Wei et al. [65] for more details. PheWAS studies analyze a wide range of phenotypes affected by a specific genetic variant. Denny et al. [71] applied computational phenotyping on EHR to automatically detect 776 different disease populations and their matched controls. Statistical tests were then carried out to determine associations between single nucleotide polymorphisms and multiple disease phenotypes. Additional studies have established the efficiency of EHR-based PheWAS to detect genetic association [72-74]. Compared to traditional genomic research, computational phenotyping has driven discovery of variant-disease associations and has facilitated the completion of genomic research in a timely and lower cost manner [66].

# 3 METHODS FOR NLP-BASED COMPUTATIONAL PHENOTYPING

NLP methods for computational phenotyping algorithms exhibit a wide range of complexities. Early stage systems were often based on keyword search or customized rules. Later, supervised statistical machine learning algorithms were applied extensively to computational phenotyping. More recently, unsupervised learning has resulted in effective patient representation learning and discovery of novel phenotypes. This section reviews NLP methods for EHR-based computational phenotyping, starting with three major categories: 1) keyword search or rule-based systems, 2) supervised learning systems, and 3) unsupervised systems. We then identify current trends and active directions of development. For convenience, we summarize the characteristics of studies reviewed in this section in

Table 1. The studies are characterized regarding the methods used to generate features, the methods or tools used for classifying the assertions (e.g., negations) of the features, the named entity recognition methods used to identify the concepts in the narratives, and the data sources used for modeling training.

## 3.1 Keyword Search and Rule-based System

Keyword search is one of the algorithms with the least model complexity for computational phenotyping. It looks for keywords, derivations of those keywords, or a combination of keywords to extract phenotypes [75]. For example, "pneumonia in the right lower lobe" is a derivation of the key phrase "consolidation in the left lower lobe" in Fiszman et al. [75]. These keywords correspond to medications, diseases, or symptoms; and, in practice, they are often identified using regular expressions. In early work, large tables of keywords were generated. Meystre et al. [76] manually built a keyword table using 80 selected concepts with related sub-concepts. They retrieved 6,928 phrases corresponding to the 80 concepts from the Unified Medical Language System (UMLS) Metathesaurus *MRCONSO* table [77]. After filtering, they still had 4,570 keywords remaining. Based on these keywords for classification, they achieved a precision of 75% and a recall of 89%. Wagholikar et al. [78] developed a keyword search system for limb abnormality identification using free-text radiology reports. Even though the reports have an average length of only 52 words, they achieved an F-measure of 80% and an accuracy of 80%. Despite their success, problems caused by the unstructured, noisy nature of the narrative text (e.g., grammatical ambiguity, synonyms, term abbreviation, misspelling, or negation of concepts) remain bottlenecks in keyword search. In general, keyword search is more susceptible to low accuracy due to simplicity of features. To improve model performance, supplementary rules (or other more sophisticated criteria) have been added to keyword search.

Rule-based systems are among the most frequently used computational phenotyping methods. In a review by Shivade et al. [3], 24 out of 97 computational phenotyping related articles have described rule-based systems. In a typical rule-based system, criteria need to be pre-defined by domain experts. For example, Wiley et al. [79] developed a rule-based system for stain-induced myotoxicity detection. They manually annotated 300 individuals' allergy listings and pre-defined a set of keywords. Then they developed a set of rules to detect contextual mentions around the keywords. In this study, they achieved a positive predictive value (PPV) score of 86% and a negative predictive value (NPV) score of 91%. Ware et al. [80] developed a list of concepts together with a list of secondary concepts that appear in the same sentence. The secondary concepts were mainly medications. After defining the concepts, they developed a set of rules for phenotype identification. This framework achieved an overall kappa score of 92% with the original annotations. Nguyen et al. [40] implemented an NLP tool, called the General Architecture for Text Engineering (GATE), to extract UMLS concepts and mapped them to SNOMED CT concepts. These SNOMED CT concepts are utilized to detect the stage of lung cancer using defined rules based on staging guidance. They achieved accuracies of 72%, 78%, and 94% for T, N, and M staging, respectively.

Xu et al. [34] implemented a heuristic rule-based approach for colorectal cancer assertion. The system used MedLEE [81] to detect colorectal cancer-related concepts. It then applied defined rules to search for concept contexts. The system achieved an F-measure of 99.6% for document level concept identification. Li et al. [82] developed a rule-based system to detect adverse drug events and medical errors using patients' clinical narratives, medications, and lab results. They compared the model's performance to a trigger tool [83], and they achieved 100% agreement. The triggers in the trigger tool are a combination of keywords that signal an underlying event of interest. Haerian et al. [84] defined rules to extract concepts from discharge summaries on top of the ICD-9 code. The use of concepts increased the model's PPV score from 55% to 97%. Sauer et al. [85] developed a set of rules to identify bronchodilator responsiveness from pulmonary function test reports, and they achieved an F-measure of 98%.

Rule-based systems often need many complex attribute-specific rules, which may be too rigid to account for the diversity of the language expression. As a result, rule-based systems may exhibit have high precision, but low recall. In fact, as will be detailed in the next subsections, more recent systems opted to use statistical machine learning algorithms to replace or complement rules.

Developing rules is laborious, time-consuming and requires expert knowledge. Despite these disadvantages, rule-based systems remain one of the most popular computational phenotyping methods in the field due to their straightforward construction, easy implementation, and high accuracy [30].

## 3.2 Supervised Statistical Machine Learning Algorithms

To improve upon accuracy and scalability while decreasing domain expert involvement, statistical machine learning methods have been adopted for computational phenotyping. These methods usually have the advantage that in addition to classifying phenotypes, they often provide the probability or confidence of that classification. In general, statistical machine learning methods are categorized as supervised, semi-supervised, or unsupervised. Common to all methods, each subject is represented as a vector consisting of features. In supervised learning, each sample in a training dataset is labeled. Algorithms predict the labels for an unknown or test dataset after learning from the training dataset. In contrast, unsupervised learning identifies patterns without labeling. It automatically clusters samples with similar patterns into groups. Semi-supervised algorithms reflect a middle ground and are used when we have both labeled and unlabeled samples. Among the most widely used supervised learning algorithms for computational phenotyping are logistic regression, Bayesian networks, support vector machines (SVMs), decision trees, and random forests. More introductory and detailed description of supervised and unsupervised methods can be found in review papers such as Kotsiantis et al. [86] and Love et al. [87].

Regression methods have a long history of application for computational phenotyping [15, 28, 29]. Regression models adjust their parameters to maximize the conditional likelihood of the data. Further, regression models do not require a lot of effort in building or tuning, and the feature statistics derived from these regression models can be easily interpreted for meaningful insights.

In a study of identification of methotrexate-induced liver toxicity in patients with rheumatoid arthritis, Lin et al. [29] collected Concept Unique Identifiers (CUIs), Methotrexate (MTX) signatures, nearby words, and part-of-speech (POS) tags as features for an L2-regularized logistic regression. They obtained an F-measure of 83% in a performance evaluation. Liao et al. [88] implemented adaptive least absolute shrinkage and selection operator (LASSO) penalized logistic regression as classification algorithm to predict patients' probabilities of having Crohn's disease and achieved a PPV score of 98%. Both Lin's and Liao's methods experimented with a combination of features from structured EHR and NLP-processed features from clinical narratives. Their studies showed that the inclusion of NLP methods resulted in significantly improved performance for regression models. Due to the high dimensionality of features extracted from narratives, both methods applied regularized regressions.

Both Naive Bayes and Bayesian network classifiers are probabilistic classifiers [89] and work well with high-dimensional features. Unlike Bayesian networks, Naive Bayes doesn't require the inference of a dependency network and is more convenient in application when feature dimension is large. This is because Naive Bayes models assume that features are independent of one another whereas Bayesian networks allow for dependency among features. Besides their simplicity, Naive Bayes models are particularly useful for large datasets and are less prone to overfitting—sometimes outperforming highly sophisticated classification methods when sufficient data are available [90]. For example, Pakhomov et al. applied Naive Bayes to predict heart failure [91], using coded data (e.g., ICD-9, SNOMED) and a "bag of words" representation from clinical narratives as features. They chose Naive Bayes for their predictive algorithm due to its ability to process high-dimensional data. Their model achieved a sensitivity of 82% and a specificity of 98%. Similarly, Chase et al. [92] applied Naive Bayes for multiple sclerosis classifications and obtained an AUC score of 90%. Some studies have suggested that results obtained from logistic regression and Naive Bayes are comparable [93]. Compared to logistic regression, the Naive Bayes classifier is capable of learning even in the presence of some missing values and relies less on missing data imputation [94, 95].

**Table 1 Summarization and characterization of computational phenotyping systems. Abbreviations: CPT Current Procedural Terminology; CUI Concept Unique Identifier; cTAKES clinical Text Analysis and Knowledge Extraction System; ICD-9 International Classification of Diseases, ninth revision; NLP Natural Language Processing; UMLS Unified Medical Language System; TF-IDF Term Frequency-Inverse Document Frequency; HITEx Health Information Text Extraction; MedLEE Medical Language Extraction and Encoding System; KMCI KnowledgeMap Concept Indexer; NILE Narrative Information Linear Extraction.**

| Study | Assertion | Concept Extraction/ Concept Mapping | Data Source | Feature Generation |
|---|---|---|---|---|
| Aramaki et al. [96] | NA | Self-defined keywords | Narrative | Similarity score between sentences |
| Bejan, Vanderwende, et al. [97] | Section headers, self-defined features, NegEx, and ConTex | MetaMap | Restricted set of time order physician daily note | Uni-grams, bi-grams, UMLS concepts, assertion values associated with pneumonia expressions, statistical significance testing to rank features |
| Carroll et al. [27] | Modified form of NegEx in KMCI, section header | KMCI, MedEx for medication | Clinical notes, ICD-9 | ICD-9, medication name, CUI, total note counts |
| Carroll et al. [35] | HITEx, Customized NegEx queries | HITEx | Diagnosis, billing, medication, procedural codes, physician text notes, discharge summaries, laboratory test results, radiology report | 21 defined attributes from patients' narrative |
| Chapman et al. [98] | NA | SymText | X-ray reports | Pneumonia-related related concepts and its states from SymText |
| Chase et al. [92] | NA | MedLEE | Narrative | 50 buckets representing pools of synonymous UMLS terms |
| Chen et al. [99] | NA | KMCI, SecTag, MedLEE, MedEx | Narrative, ICD-9, CPT | ICD-9, CPT, CUIs |
| Castro et al. [100] | Context dependent tokenizer in cTAKES | cTAKES | Radiology reports | Concepts, context dependent concepts, and concepts from cTAKES |
| Davis et al. [30] | Negation, word-sense disambiguation tool in KMCI | KMCI | ICD-9 codes, free text, and medications | ICD-9, CUIs, keywords |
| DeLisle et al. [101] | Customized rules, NegEx | Examined UMLS-supplied lexical variants/semantic types | Narrative, ICD-9, vital signs and orders for tests, imaging, and medications | 186 UMLS associated with phenotype |
| DeLisle et al. [102] | NA | cTAKES | Chest imaging report ICD-9, encounter information, prescriptions | ICD-9, antibiotics medicine, hospital re-admission, binary variable of non-negative of chest imaging report |
| Fiszman et al. [75] | Self-defined rules | SymText | Chest x-ray reports | Set of augmented transition network grammars and a lexicon derived from the specialist Lexicon |
| Garla et al. [103] | cTAKES, YTEX, defined rules | Use cTAKES and YTEX to map concepts to UMLS and customized dictionary | Narrative and customized dictionary | Terms suggestive of benign/malignant lesions and UMLS concept in any liver-cancer related sentence |
| Gehrmann et al. [104] | cTAKES | cTAKES | Discharge summary | Concepts from cTAKES were transformed to continuous features using the TF-IDF |

| Haerian and Salmasian et al. [84] | Manual, and MedLEE | MedLEE map concepts to UMLS | Discharge summaries, ICD-9 | MedLEE concepts were manually reviewed by a clinician. 31 codes were used |
|---|---|---|---|---|
| Herskovic et al. [105] | NA | MetaMap, SemRep | Narrative, biomedical literature | UMLS concept and UMLS relationship, semantic predication from biomedical literature |
| Lehman et al. [106] | NegEx | map to customized UMLS dictionary | Narrative | Manually selected UMLS concept |
| Li et al. [82] | NA | NA | Narrative, medication, lab results | Neonatologists manually reviewed 11 patients' notes and defined keywords and rules |
| Liao et al. [15] | Occurrence of concepts to indicate positive or negative of a sentence | HITEx | Provider notes, radiology reports, pathology reports, discharge summaries, operative reports, ICD-9, prescriptions | Concepts from HITEx, count of the concepts, binary variable to indicate occurrence of concepts. |
| Liao et al. [107] | NA | HITEx | ICD-9, CPT, lab results, narrative | Binary variable was created to indicate whether a concept was mentioned or not |
| Lin et al. [29] | cTAKES | cTAKES | Narrative, medication code, customized CUI | CUI, drug signatures (dosage, frequency), temporal features, nearby words, nearby POS tags |
| Luo et al. [108] | NA | Stanford Parser, Link Parser, ClearParser | Narrative | CUIs were used as nodes in the graph, syntactic dependencies among the concepts were used as edges in the graph |
| McCowan et al. [109] | NegEx | UMLS mapper | Pathology report | Map UMLS concepts to specific factors from the staging guidelines |
| Nguyen et al. [40] | NegEx, section heading | MEDTEX | Narrative | Concepts related to lung cancer resections (based on the AJCC 6th edition) were used |
| Ni et al. [54] | NegEx | cTAKES map to UMLS, SNOMED CT | Encounter data and clinical notes | Use concepts and encounter data. Predefine concepts from selection criteria, search for the hyponyms of query word |
| Nunes et al. [110] | Manual | NA | Narrative, ICD-9, lab results, demographics | Manual extract related terms and hyponyms and related words |
| Peissig et al. [111] | MedLEE | MedLEE | Narrative, ICD-9, CPT | UMLS concepts, ICD-9, CPT |
| Pineda et al. [112] | ConText | Topaz pipeline, map to UMLS | Narrative, lab test | Selected UMLS concepts and two lab test concepts |
| Posada et al. [113] | Section titles | MedLEE, keyword extraction, Question-Answer Feature Extraction | Psychiatric evaluation records | Count of keywords and concepts fall in nine defined categories as feature |
| Roque et al. [114] | NA | Simple sentence splitter split the text into smaller units | ICD-10, narrative | ICD-10, small units of sentences |
| Sauer et al. [85] | NA | Manual | Narrative | Experts reviewed notes and collected patterns to design |

| | | | | extraction rules using regular expression |
|---|---|---|---|---|
| South et al. [115] | Negex | UMLS Metathesaurus | Narrative | NA |
| Teixeira et al. [116] | NegEx | MetaMap | Narrative, document count, medication, hypertension lab test related structured data | UMLS concepts, SNOMED-CT generated from narrative, ICD-9 code from structured data |
| Wang et al. [117] | NA | Standford parser | Clinical notes, comments, structured files | Constituent and dependency parsed from sentence |
| Ware et al. [80] | Self-Dev | NA | Narrative | Medication, treatment, word bigrams, numerical features, synonym list |
| Wei et al. [118] | cTAKES | cTAKES, map to SNOMED-CT | Narrative | SNOMED-CT concept, semantic type, node collapse concept |
| Wilke et al. [36] | NA | FreePharma | Narrative, ICD-9, laboratory data | NA |
| Xu et al. [34] | MedLEE | MedLEE, map to UMLS CUI | Narrative, ICD-9, CPT | UMLS concept, words of distance and direction (left vs. right) |
| Yu et al. [28] | NA | NILE, map to UMLS concept | ICD-9, Narrative | ICD-9, NLP features (counts of generic drug concept), number of notes for each patient |
| Zeng et al. [89] | HITEx (NexEx-2) | HITEx, map to UMLS | Narrative, ICD-9 | N-word text fragments along with frequency, UMLS concept, smoking related sentences |
| Zhao et al. [119] | Self-defined | NA | PubMed knowledge, ICD-9, narrative | Selected concepts that are associated with pancreatic cancer |

A Bayesian network consists of a directed acyclic graph whose node set contains random variables and whose edges represent relationships among the variables, and a conditional probability distribution of each node given each combination of values of its parents [120]. Bayesian networks have been used for reasoning in the presence of uncertainty and machine learning in many domains including biomedical informatics [121]. Chapman et al. [98] applied a Bayesian network inference model to predict pancreatic cancer using X-ray reports. In their experiments, a Bayesian network demonstrated high sensitivity 90% and specificity of 78%. Zhao et al. [119] applied a similar approach to identify pancreatic cancer. They developed a weighted Bayesian network with weights assigned to each node (feature). They also incorporated external knowledge from PubMed for scaling weights. Associations between each risk factor and pancreatic cancer were established using the output of NLP tools run on PubMed. Finally, they selected 20 risk factors as variables and fit them into a weighted Bayesian network model for pancreatic cancer prediction. Their results showed that this weighted Bayesian network achieved an AUC score of 91%, which had better performance than a traditional Bayesian network (81%). Compared to logistic regression or Naive Bayes methods, as a probabilistic formalism, Bayesian networks offer a better capacity to integrate heterogeneous knowledge in a single representation, which is particularly important in computational phenotyping because it complements the increasing availability of heterogeneous data sources [119]. A priori estimations can be taken into account in Bayesian network; this advantage allows one to incorporate known domain knowledge to increase model performances.

Clinical narratives are known to have high-dimensional feature spaces, few irrelevant features, and sparse instance vectors [122]. These problems were found to be well-addressed by SVMs [122]. In addition, SVMs have been recognized for their generalizability and are widely used for computational phenotyping [27, 89, 97, 103, 109, 123, 124]. In SVM models, a classifier is created by maximizing the margin between positive and negative examples [125]. Wei et al. [118] applied Mayo clinical Text Analysis and Knowledge Extraction System (cTAKES) to extract SNOMED CT concepts from clinical documents. The concepts were used to train a SVM for Type 2 Diabetes identification. Their algorithm achieved an F-measure of 95%. They concluded that concepts from the semantic type of disease or syndrome contain most important information for accurate phenotyping. Carroll et al. [27] implemented a SVM model for rheumatoid arthritis identification using a set of features from clinical narratives using the Knowledge Map Concept Identifier (KMCI) [126]. They demonstrated that a SVM algorithm trained on these features outperformed a deterministic algorithm. Zeng et al. [89] trained a SVM model for principal diagnosis, co-morbidity, and smoking status identification. The features for the model were concepts extracted from discharge summaries and ICD-9 codes. The model achieved accuracies of 90% for smoking status, 87% for co-morbidity, and 82% for principal diagnoses. Chen et al. [99] applied active learning to a SVM classification algorithm to identify rheumatoid, colorectal cancer, and venous thromboembolism. Their results showed that active learning with a SVM could reduce annotated sample size while remaining relatively high performance. In the reviewed papers, SVMs constantly outperform other learning algorithms for computational phenotyping [27, 89, 99, 118, 127].

Kernel methods provide a structured way to extend the use of a linear algorithm to data that are not linearly separable by transforming the underlying feature space. The nonlinear transformation enables it to operate on high-dimensional data without explicitly computing the coordinates of the data in that space. SVMs are the most well-known learning algorithm using kernel based methods. Kotfila et al. [128] evaluated different SVM kernels' performances in identifying five diseases from unstructured medical notes. They found that SVMs with Gaussian radial basis function (RBF) kernels outperformed linear kernels. Zheng et al. [129] found that a SVM with RBF kernel exceeded non-kernel-based SVMs, decision trees, and perceptron for coreference resolution identification from the clinical narrative. In a study by Turner et al. [130], the authors tried to identify Systemic Lupus Erythematosus (SLE) from clinical notes. The authors concluded that a SVM with linear kernel outperformed radial basis function, polynomial, and sigmoid kernels. Good performance can be achieved in kernel methods with the appliance of statistical learning theory or Bayesian arguments. Linear methods are favored when there are many samples in a high dimensional input space. In contrast, for low-dimensional problems with many training instances, nonlinear kernel methods may be more favorable. Apart from the models mentioned above, researchers have explored other methods such as random forests [112], decision trees [100, 113, 131, 132], and the Longitudinal Gamma Poisson Shrinker [133, 134] for computational phenotyping. DeLisle et al. [102] implemented a conditional random field probabilistic classifier [135] to identify acute respiratory infections. They used structured data combined with narrative reports and demonstrated the inclusion of free text improved the PPV score by 20–70% while retaining sensitivities around 58-75%. Chapman et al. [98] applied decision trees, Bayesian networks, and an expert-crafted rule-based system to extract bacterial pneumonia from X-ray reports. The method using decision trees achieved an AUC score of 94%, and it is close to the other systems. Furthermore, semi-supervised methods have also been investigated for computational phenotyping [136, 137], which have the potential to significantly reduce the amount of labeling work and simultaneously retain high accuracy. Aramaki [96] applied K-Nearest Neighbor classifier [138] based on the Okapi-BM25 similarity measure to extract patient smoking statuses from free text, and they achieved 89% accuracy in a performance evaluation. Carrero et al. [139] applied AdaBoost with Naive Bayes for text classification, and they achieved an F-measure of 72% using bigrams. Ni et al. [54] used TF-IDF similarity scores calculated from the feature vectors to identify a cohort of patients for clinical trial eligibility prescreening. Hybrid methods make use of more than one methods have also received increasing attention [138, 139], suggesting a promising direction for practical performance improvement.

For many data resources and domains, various models have been investigated, and some of them have achieved impressive success. However, a comprehensive understanding of the superior performance of a particular method over another for a specific domain remains an open challenge.

## 3.3 Unsupervised Learning

The time-consuming and labor-intensive process of obtaining labels for supervised learning algorithms limits their applicability to computational phenotyping. Another limitation of supervised learning is that it only looks for known characteristic patterns by designating a task and its outcome [86]. Unsupervised learning, on the other hand, can automatically classify phenotypes without extra annotations by experts [105, 140, 141]. Moreover, unsupervised learning searches for intrinsic patterns of data. Luo et al. [142] introduced subgraph augmented non-negative tensor factorization (SANTF) to cluster patients with lymphomas into three subtypes. After extracting atomic features (i.e., words) from narrative text, they implemented SANTF to mine relation features to cluster patients automatically. Their study demonstrated that NLP methods for unsupervised learning were able to achieve a decent accuracy (75%) and at the same time to discover latent subgroups. Roque et al. [114] extracted concepts from free text and mapped them to ICD-10 code. The ICD-10 code vector was used to represent each patient's profile and cosine similarity scores between each pair of ICD-10 vectors were obtained. Then, they applied hierarchical clustering to cluster those patients based on cosine similarity scores. As a result, they identified 26 clusters within 2,584 patients. They further analyzed the clinical characteristics of each cluster and concluded that NLP-based unsupervised learning was able to uncover the latent pattern of patient cohorts. Ho et al. [143] applied sparse non-negative tensor factorization on counts of normal and abnormal measurements obtained from EHR data for phenotype discovery. They identified multiple interpretable and concise phenotypes from a diverse EHR population, concluding that their methods were capable of characterizing and predicting a large number of diseases without supervision. Quan et al. [144] applied kernel-based pattern clustering and sentence parsing for interaction identification from narratives. In their application of protein-protein interaction, the unsupervised system achieved close performance to supervised methods.

Unsupervised learning has mitigated the laborious labeling work, thus making studies more scalable, and has the capability of finding novel phenotypes. However, interpretation of these new phenotypes requires domain expertise and remains challenging. Additionally, model performance in unsupervised learning is not yet as good as supervised learning. EHR-based unsupervised learning has frequently been applied on structured data [44, 45], but less frequently on narratives [142]. Further investigations on incorporating multiple data sources and at the same time maintaining or improving the performance are expected.

## 3.4 Deep Learning

Deep learning algorithms are good at finding intricate structures in high-dimensional data and have demonstrated good performance in natural language [145]. They have been adapted to learn vector representations of words for NLP-based phenotyping [112, 136], laying a foundation for computational phenotyping. Deep learning has been applied on various NLP applications, including semantic representation [146], semantic analysis [147, 148], information retrieval [149, 150], entity recognition [151, 152], relation extraction [153-156], and event detection [157, 158].

Beaulieu-Jones et al. [136] developed a neural network approach to construct phenotypes to classify patient disease status. The model obtained better performance than SVM, random forest, and decision tree models. They also claimed to successfully learn the structure of high-dimensional EHR data for phenotype stratification. Gehrmann et al. [104] compared convolutional neural networks (CNNs) to the traditional rule-based entity extraction systems using the cTAKES and logistic regression using n-gram features. They tested ten different phenotyping tasks using discharge summaries. The CNNs outperformed other phenotyping algorithms in the prediction of ten phenotypes, and they concluded that NLP-based deep learning methods improved the performance of patient phenotyping compared to other methods. Wu et al. [159] applied CNNs using a set of pre-trained embeddings on clinical text for named entity recognization. They found that their models outperformed the baseline of conditional random fields (CRF). Geraci et al. [160] applied deep neural networks to identify youth depression from unstructured text notes. The authors achieved a sensitivity of 93.5% and a specificity of 68%. Jagannatha et al. [161, 162] experimented with recurrent neural networks (RNNs), long short-term memory (LSTM), gated recurrent units (GRUs), bidirectional LSTMs, combinations of LSTMs with CRF, and CRF to extract clinical concepts from texts. They found that all variants of RNNs outperformed the CRF baseline. Lipton et al. [163] evaluated the performance of LSTM in phenotype prediction using multivariate time series clinical measurements. They concluded that their model outperformed logistic regression and multi-layer perceptron (MLP). They also concluded that the combination of LSTM and MLP had the best performance. Che et al. [164] also applied deep learning methods to study time series in ICU data. They introduced a prior-based Laplacian regularization process on the sigmoid layer that is based on medical ontologies and other structured knowledge. In addition, they developed an incremental training procedure to iteratively add neurons to the hidden layer. Then they applied causal inference techniques to analyze and interpret the hidden layer representations. They demonstrated that their proposed methods improved the performance of phenotype identification and that the model trains with faster convergence and better interpretation.

It is commonly known that unsupervised pre-training can improve deep learning performances and generalizability [165]. A generative deep learning algorithm that uses unsupervised methods can be applied to large unlabeled datasets, which has the potential to increases model generalizability [166]. Miotto et al. [167] applied a deep learning model called an auto-encoder as an unsupervised model to learn the latent representations for patients in order to predict their outcome and achieved better performance than principal component analysis. Due to the excellent model performance and good generalizability [168], using deep learning methods in conjunction with unsupervised methods is a promising approach in NLP-based computational phenotyping. Miotto et al. [169] introduced the framework of "deep patient". The method captures hierarchical regularities and dependencies in the data to create a vector for patient representation. This study showed that pre-processing data using a deep sequence of non-linear transformations can help better information embedding and information inference. Word2Vec [170] is an unsupervised artificial neural network (ANN) that has been developed to obtain vector representations of words when given large corpus and the representations are dependent on the context. For more details, we refer readers to a review [16] in recent advances on deep learning techniques for EHR analysis.

Even though deep learning methods present an opportunity to build phenotyping systems with good generalizability [171], a drawback of deep learning methods is their lack of interpretability. It can be difficult to understand how the features of the model arrive at predictions even though they can train a classifier with good performance [172].

## 4 MAKING NLP MORE EFFECTIVE

With numerous NLP methods available for computational phenotyping, it is practical to consider how to select more effective NLP methods or improve current NLP methods based on problem characteristics. This section reviews existing effort in these directions including model comparison, multi-modality data integration, entity recognition, and feature relation extractions.

## 4.1 Comparison of Models

Different computational phenotyping models vary in prediction accuracy and model generalizability. Comparison studies have been carried out to explore model performances. These comparison studies indicate algorithm performance differs based on specific conditions such as data sources, features, training data sizes, and target phenotypes.

In 1999, Wilcox et al. [173] conducted a study to investigate different algorithms' performances to extract clinical conditions from narratives. These algorithms were Naive Bayes, decision table, instance-based inducer, decision tree inducer MC4, decision tree inducer C5.0, and rule-discovery inducer CN2. Outputs of NLP algorithms were used as model features. They found MC4 and CN2 had the best performances while decision table performed the worst. Chapman et al. [98] tested rule-based method, Bayesian network, and decision tree for pneumonia detection using X-ray reports. The study showed that rule-based methods had slightly better performance (AUC score: 96%) than decision tree systems (AUC score: 94%) and Bayesian networks (AUC score: 95%).

Teixeira et al. [116] found random forests were superior to rule-based systems with a median AUC score of 98% when they were trying to identify hypertension using billing codes, medications, vitals, and concepts extracted from narratives. Pineda et al. [112] compared a Bayesian network classifier, Naive Bayes, a Bayesian network with the K2 algorithm, logistic regression, neural network, SVM, decision tree, and random forest for influenza detection. They concluded that all the machine learning classifiers had good performance with AUC score ranging from 88% to 93% and outperformed curated Bayesian network classifier, which had an AUC score of 80%.

Dumais et al. [132] compared the performances of SVM, Naive Bayes, Bayesian networks, decision trees, and rule-based systems in text classification. They concluded SVMs showed the best performance and noted that the training process is fast. Chen et al. [99] applied active learning to SVM classification, and their results showed that active learning with a SVM could reduce sample size needed. They concluded that semi-supervised learning, such as active learning, is efficient insofar as it reduces labeling cost.

Gehrmann et al. [104] compared convolutional neural networks (CNNs) to logistic regression and random forest model. They found CNNs had an improved performance compared to others and it can automatically learn the phrases associated with each patient phenotype, which reduced annotation complexity for clinical domain experts.

Among the compared methods, keyword search and rule-based systems often achieve good performance when such systems are well-designed and well-tuned. However, the construction of a keyword and rule list is laborious, making these systems difficult to scale. Supervised machine learning models have been favored for their capabilities of acquiring classification patterns and structures from data. The performance of supervised methods varies depending on the sample size, data resource type, number of data resources. Deep learning has also been favored for its better performance and generalizability. It has also been suggested that inclusion of more data resources can improve the model performances [174].

## 4.2 Combining Multiple Data Modalities

Computational phenotyping often involves multiple heterogeneous data sources in addition to structured data, such as clinical narratives, public databases, social media, biomedical literature [15, 88, 101, 111, 115, 175, 176]. Adding heterogeneous data has the benefit of providing complementary perspectives for computational phenotyping models [117]. Teixeira et al. [116] tested different combinations of ICD-9 codes, medications, vitals, and narrative documents as data resources for hypertension prediction. They found that model performance increases with the number of data resources regardless of the method used. They concluded that combination of multiple categories of information result in the best performances. The complete list of data sources utilized in the reviewed literature appears in

Table 1.

Liao et al. [15, 107] compared algorithms using ICD-9 codes alone to algorithms using a combination of structured data and NLP features. The results showed that the incorporation of NLP features improved algorithm performance significantly. Similarly, Nunes et al. [110] concluded that both structured data and clinical notes need to be considered to assess the occurrence of hypoglycemia among diabetes patients fully. Yu et al. [28] collected concepts from publicly available knowledge sources (e.g., Medscape, Wikipedia) and combined them with concepts extracted from narratives to predict rheumatoid arthritis (RA) and coronary artery (CAD) disease status. Their results showed that the combination of available public databases like Wikipedia and features derived from narratives could achieve high accuracy in RA and CAD prediction. Xu et al. [34] used ICD-9 codes, Current Procedural Terminology (CPT) codes, and colorectal cancer concepts to identify colorectal cancer. Zhao et al. [119] applied additional PubMed knowledge to weight the existing features.

The increasing trend of combining multiple data sources reflects the increased availability of EHR data and publicly available data [26]. Also, coupled with the increasing model complexities, there is a potential that more comprehensive data sources will be included for computational phenotyping. For example, one application developed by Gehrmann et al. [104] used CNNs to automatically learn the phrases associated with patients' phenotypes without task-specific rules or pre-defined keywords, which reduced the annotation effort for domain experts. As such, various data sources can be adopted for model training without too much human labor. However, regarding model generalizability, models and features based on narratives do not appear to be as portable as the ones based on structured EHR fields [116].

## 4.3 Entity Recognition and Relation Extraction

It is important to accurately recognize entities in clinical narratives as the extracted concepts are often used as features for models. Methods for feature learning vary from early-on manual selection to, more recently, machine learning methods. State-of-the-art named entity recognizers can automatically annotate text with high accuracy [177]. Bejan et al. [97, 123] implemented statistical feature selection, such as logistic regression with backward elimination to reduce feature dimensions. Wilcox et al. [173] tested machine learning algorithms with both expert-selected variables and automatically-selected variables by identifying top ranking predictive accuracy variables to classify six different diseases. Several studies, including those of Lehman et al., Luo et al., and Ghassemi et al. [106, 142, 178, 179], applied topic models and extended tensor-based topic models to learn better coherent features. Chen et al. [180] have applied an unsupervised system that is based on phrase chunking and distortional semantics to find features that are important to individual patients. Zhang et al. [181] have applied an unsupervised approach to extract named entities from biomedical text. Their model is a stepwise method, detecting entity boundaries and also classifying entities without pre-defined rules or annotated data. To do this, they assume that entities of same class tend to have similar vocabulary and context, which is called distributional semantics. Their model achieves a stable and competitive performance.

In addition to features, it is also critical to capture relations among features. Understanding these relations is important for knowledge representation and inference to augment structured knowledge bases [182, 183]. To date, a majority of the state-of-the-art methods for relation extraction are graph-based. Xu et al. [184] developed medication information extraction system (MedEx) to extract medications and relations between them. They applied the Kay Chart Parser [185] to parse sentences according to a self-defined grammar. In this way, they converted narratives to conceptual graph representations of medication relations. Using this graph representation, they were able to extract the association strength, frequencies, and routes. Representing medical concepts with graph nodes, Luo et al. [108]

augmented the Stanford Parser with UMLS-based concept recognition to generate graph representations for sentences in pathology reports. They then applied frequent subgraph mining to collect important semantic relations between medical concepts. The integration of named entity detection with relation extraction will produce end-to-end systems that can further automate the discovery and curation of novel biomedical knowledge. In addition, there is a trend towards increasingly unsupervised relation extraction, which is more adaptable across biomedical subdomains. Unsupervised methods have been investigated for feature relations too. Ciaramita et al. [186] presented an unsupervised model to learn semantic relations from text, hypothesizing that semantically related words co-occur more frequently. The model represented relations as syntactic dependency paths between ordered pairs of named entities. Relations were selected using the similarity scores associated with each class pair and dependency paths. Most recently, Alicante et al. [187] proposed using unsupervised methods for both entity and relation extraction from clinical notes. Clustering was applied to all the entity pairs for possible relations discovery.

# 5 FUTURE WORK

While notable progress has been made in computational phenotyping, challenges remain in developing generalizable, efficient, and effective models for accurate phenotype identification. Below we discuss these challenges and directions for future work.

## 5.1 Information heterogeneity in clinical narratives

Boland et al. [188] highlighted the heterogeneity apparent in clinical narratives due to the variance in physicians' expertise and behaviors. Different clinicians' perspectives can be quite different, and in practice they often are. Also, clinical narratives are often ungrammatical, incomplete with limited context, and contain a large number of abbreviations and acronyms [189], all of which make computational phenotyping challenging. Studies have applied UMLS or other external controlled vocabularies to recognize the various expressions of the same medical concept. However, performances of those external modules remain controversial [190, 191]. How to resolve the heterogeneity in clinical narratives remains an interesting topic.

## 5.2 Model generalizability

There is an ongoing trend of expanding generalizable algorithms to mine multiple diseases from different narratives. But these methods are still lacking in computational phenotyping [192, 193]. In addition, rule-based systems are one of the most prevalent methods for NLP-based computational phenotyping [3]. The intensive human labor required to adapt rules to a new system affects the model generalizability. Studies investigating algorithms that automatically mine rules are not yet available. Furthermore, even though statistical analysis and machine learning have provided alternative ways to automatically generate phenotypes, high dimensional feature spaces, data sparsity, and data imbalance remain impediments to the adoption of these methods [194]. Development of complete pipelines using various data sources for different phenotypes is one potential solution for generalizable computational phenotyping.

## 5.3 Model interpretability

More sophisticated models, such as convolutional neural networks, have the potential to automatically learn the phrases associated with each phenotype, which can reduce annotation complexity for clinical domain experts [104]. Using such models, one might be able to develop a system with good generalizability and have the availability to use multiple data sources. However, these same models tend to lack interpretability, which presents a problem that remains to be solved. Furthermore, meaningful interpretations of the novel phenotypes discovered in unsupervised clustering models remain one of the next big challenges in the field. Another promising direction is improving interpretation *while* retaining, or even improving, performance.

## 5.4 Characterizing the context of computational phenotyping

Clinical narratives contain patients' concerns, clinicians' assumptions, and patients' past medical histories. Clinicians also record diagnoses that are ruled out or symptoms that patients denied. Conditions, mentions, and feature relations can be extracted to better distinguish differential diagnoses. In computation phenotyping, generalized relation and event extraction, rather than binary relation classification, are expected to be a promising direction for future research; especially for the tasks of extracting clinical trial eligibility criteria [195], representing test results for automating diagnosis categorization [108], and building pharmacogenomic semantic networks [58], where the number of nodes is flexible, and the relation structure may not be entirely pre-specified due to the high complexity. To this end, graph methods are a promising class of algorithms and should be actively investigated [108, 142].

# 6 CONCLUSION

In this paper, we review the applications of NLP methods for EHR-based computational phenotyping, including the state-of-the-art NLP algorithms for this task. Our review shows that the keyword search, rule-based methods, and supervised machine learning-based NLP are the most widely used methods in the field. Well-designed keyword search and rule-based systems often show high accuracy. However, manually constructing keyword lists and rules results in problematically low generalizability and scalability for those methods.

Supervised classification has higher accuracy and is easy to train and test. However, the supervised classification methods require the training samples to be labeled, which can be labor intensive. To date, there is not a dominating method in the field; rather, model performances for the same type of methods may even vary depending on the data sources, data types, and sample sizes.

The combination of different data sources has the potential to improve model performance. Recently, unsupervised machine learning algorithms are gaining more attention because they require less human annotation and hold potential for finding novel phenotypes. Furthermore, new developments in machine learning methods, such as deep learning, have been increasingly adopted.

Finally, there is an emerging trend to extract relations between medical concepts as more expressive and powerful features. The extracted relations have been shown to increase algorithm performance significantly.

Despite these advances across multiple frontiers, there are many remaining challenges and opportunities for NLP-based computational phenotyping. These challenges include better model interpretability and generalizability, as well as proper characterization of feature relations in clinical narratives. These challenges will continuously shape the emerging landscape and provide research opportunities for NLP methods in EHR-based computational phenotyping.

## ACKNOWLEDGMENT

This work was supported in part by NIH Grant 1R21LM012618-01, NLM Biomedical Informatics Training Grant 2T15 LM007092-22, and the Intel Science and Technology Center for Big Data. Corresponding author: Yuan Luo (yuan.luo@northwestern.edu)

Zexian Zeng received both his master degrees in Industrial and Systems Engineering and Computer Science from University of Wisconsin-Madison in 2014. Currently, he is working towards the Ph.D. degree at the Department of Preventive Medicine, Northwestern University Feinberg School of Medicine. His research interests are in natural language processing and cancer genomics.

Yu Deng earned her bachelor degree in Biotechnology from Northeast Normal University in 2014, China. Currently, she is working towards her PhD degree in Biomedical Informatics, Northwestern University. She works on the development of mathematics and computer methods for dynamic risk prediction. She is also interested in disease sub-phenotype discovery using clustering methods. She is a member of American Medical Informatics Association since 2016.

Xiaoyu Li received a BA in Sociology from Tsinghua University in 2011, an MS in sociology from University of Wisconsin-Madison in 2013, and an MS in Biostatistics and a ScD in Social and Behavioral Sciences from Harvard T.H. Chan School of Public Health in 2017. She is currently a postdoctoral research fellow at Harvard T.H. Chan School of Public Health and Brigham and Women's Hospital. She is interested in data science methods and contextual determinants of population health. She is a reviewer for the journal of Social Science and Medicine and a member of the Society for Epidemiologic Research.

Tristan Naumann is a Ph.D. candidate in Electrical Engineering and Computer Science at MIT working with Professor Peter Szolovits in CSAIL's Clinical Decision-Making group. His research includes exploring relationships in complex, unstructured healthcare data using natural language processing and unsupervised learning techniques. He has been an organizer for workshops and datathon events, which bring together participants with diverse backgrounds in order to address biomedical and clinical questions in a manner that is reliable and reproducible.

Yuan Luo is an assistant professor in the Department of Preventive Medicine at Northwestern University Feinberg School of Medicine. He received his PhD in Computer Science from Massachusetts Institute of Technology. He served on the student editorial board of Journal of American Medical Informatics Association. His research interests include machine learning, natural language processing, time series analysis, computational genomics, with a focus on biomedical applications. Dr. Luo is the recipient of the inaugural Doctoral Dissertation Award Honorable Mention by American Medical Informatics Association (AMIA) in 2017.